\documentclass[11pt,a4paper]{article}
\usepackage[acceptedWithA]{acl2020}
\usepackage{times}
\usepackage{latexsym}

\usepackage{natbib}
\usepackage{graphicx}
\usepackage{todonotes}
\usepackage{times}
\usepackage{latexsym}
\usepackage{color} 
\usepackage{graphicx}
\usepackage{stfloats}
\usepackage{booktabs}
\usepackage{array}
\usepackage{amssymb, amsmath}
\usepackage{multicol}
\usepackage{multirow}
\usepackage{threeparttable}
\usepackage{algorithm}
\usepackage{algorithmic}

\usepackage{arydshln}

\usepackage{enumitem}
\usepackage{makecell}
\usepackage{xcolor}
\definecolor{mygreen}{HTML}{3cb44b}
\usepackage{microtype}

\usepackage{amsmath}
\usepackage{amsfonts}
\usepackage{amssymb}
\usepackage{wrapfig}
\usepackage{subcaption}
\usepackage{multirow}
 \usepackage{mathtools} 

\usepackage{verbatim}

\usepackage{anyfontsize}

\usepackage{color}
\usepackage{tikz}
\usetikzlibrary{arrows,shapes,snakes,automata,backgrounds,fit,petri}
\usepackage{adjustbox}

\newcommand{\RN}[1]{%
	\textup{\lowercase\expandafter{\it \romannumeral#1}}%
}

\usepackage{booktabs}

\usepackage{tcolorbox}

\makeatletter
\newcommand{\distas}[1]{\mathbin{\overset{#1}{\kern\z@\sim}}}%

\usepackage{enumitem}

\newcommand{\ie}[0]{\emph{i.e., }}

\newcommand{\eg}[0]{\emph{e.g., }}

\newcommand{\etc}[0]{\emph{etc.}}

\newcommand{\beq}{\vspace{0mm}\begin{equation}}
\newcommand{\eeq}{\vspace{0mm}\end{equation}}
\newcommand{\beqs}{\vspace{0mm}\begin{eqnarray}}
\newcommand{\eeqs}{\vspace{0mm}\end{eqnarray}}
\newcommand{\barr}{\begin{array}}
\newcommand{\earr}{\end{array}}

\newcommand{\Imat}{{\bf I}}

\newcommand{\xv}{\boldsymbol{x}}

\newcommand{\thetav}{\boldsymbol{\theta}}

\newcommand{\Lcal}{\mathcal{L}}

\newcommand{\Acal}{\mathcal{A}}

\newcommand{\Dcal}{\mathcal{D}}

\ifx\assumption\undefined
\newtheorem{assumption}{Assumption}
\fi

\ifx\definition\undefined
\newtheorem{definition}{Definition}
\fi

\ifx\remark\undefined
\newtheorem{remark}{Remark}
\fi

\usepackage{color, colortbl}
\definecolor{Gray}{gray}{0.93}

\newcommand{\data}{\textsc{FewShotWOZ}}

    \title{Few-shot Natural Language Generation for Task-Oriented Dialog}

\author{Baolin Peng, Chenguang Zhu, Chunyuan Li \\ \textbf{Xiujun Li, Jinchao Li, Michael Zeng, Jianfeng Gao} \\
  Microsoft Research, Redmond \\
  \texttt{\{bapeng,chezhu,chunyl,xiul,jincli,nzeng,jfgao\}@microsoft.com}
 }

\date{}

\begin{document}
\maketitle
\begin{abstract}
As a crucial component in task-oriented dialog systems, the Natural Language Generation (NLG) module converts a dialog act represented in a semantic form into a response in natural language. The success of traditional template-based or statistical models typically relies on heavily annotated data, which is infeasible for new domains.
Therefore, it is pivotal for an NLG system to generalize well with limited labelled data in real applications. To this end, we present \data{}, the first NLG benchmark to simulate the few-shot learning setting in task-oriented dialog systems. Further, we develop the SC-GPT\footnote{{\bf S}emantically-{\bf C}onditioned {\bf G}enerative {\bf P}re-{\bf T}raining} model. It is pre-trained on a large set of annotated NLG corpus to acquire the controllable generation ability, and fine-tuned with only a few domain-specific labels 
to adapt to 
new domains. 
Experiments on \data{} and the large Multi-Domain-WOZ datasets show that the proposed SC-GPT significantly outperforms existing methods, measured by various automatic metrics and human evaluations.
\end{abstract}


\section{Introduction}

Task-oriented dialog systems are becoming increasingly popular, as they can assist users in various daily activities such as ticket booking and restaurant reservations. In a typical task-oriented dialog system, the {\it Natural Language Generation} (NLG) module plays a crucial role: it converts a system action (\eg often specified in a semantic form selected by a dialog policy) into a final response in natural language. 
Hence, the response should be 
{\it adequate} to represent semantic dialog actions, and 
{\it fluent} to engage users' attention. As the ultimate interface to interacts with users, NLG plays a significant impact on the users' experience.


Existing methods for NLG can be broadly summarized into two major categories.
$(\RN{1})$ {\it Template-based methods} require domain experts to handcraft templates for each domain, and the system fills in slot-values afterward \citep{rule,halogen}. Thus, the produced responses are often adequate to contain the required semantic information, but not always fluent and nature, hurting users' experiences.
$(\RN{2})$ {\it Statistical language models} such as neural networks \citep{gao2019neural} learn to generate fluent responses via training from labelled corpus. One canonical model is {\it semantically conditioned LSTM} (SC-LSTM) \citep{wen-etal-2015-semantically}, which encodes dialog acts with one-hot representations and uses it as an extra feature to inform the sentence generation process. Despite its good performance on simple domains, it requires large amounts of domain-specific annotated data which is not available for many domains in real-world applications. Even worse, this renders severe scalability issues when the number of possible combinations of dialog acts grows exponentially with the number of slots in more complex domains.

We revisit the current research benchmarks for NLG, and notice that each dialog domain is extensively labelled to favor model training. However, this is in contrast to the real-world application scenarios, where only very limited amounts of labelled data are available for new domains. To simulate such a few-shot learning setting, we have developed a new benchmark dataset, called \data{}, based on the MultiWOZ \citep{budzianowski2018multiwoz} and Cambridge NLG datasets \citep{rnnlg}. 
\data{} consists of dialog utterances from 7 domains. For each domain, we provide less than 50 labeled utterances for fine-tuning.  We believe that \data{} can better inspire  research to address the challenge of learning  data-hungry statistical models with very limited amounts of labelled data in real-world scenarios.

To deal with the challenge of few-shot learning, we develop the SC-GPT model.  
SC-GPT is a multi-layer Transformer neural language model, trained in three steps:
$(\RN{1})$ Pre-trained on plain text, similar to GPT-2 \citep{gpt2}; 
$(\RN{2})$ Continuously pre-trained on large amounts of dialog-act labeled utterances corpora to acquire the ability of controllable generation;
$(\RN{3})$ Fine-tuned for a target domain using very limited amounts of domain labels.
Unlike GPT-2, SC-GPT generates semantically controlled responses that are conditioned on the given semantic form, similar to SC-LSTM but requiring much less domain labels to generalize to new domains. 






\begin{figure*}[t!]
\centering
\begin{tabular}{c c}
	\hspace{-2mm}
	\includegraphics[height=3.0cm]{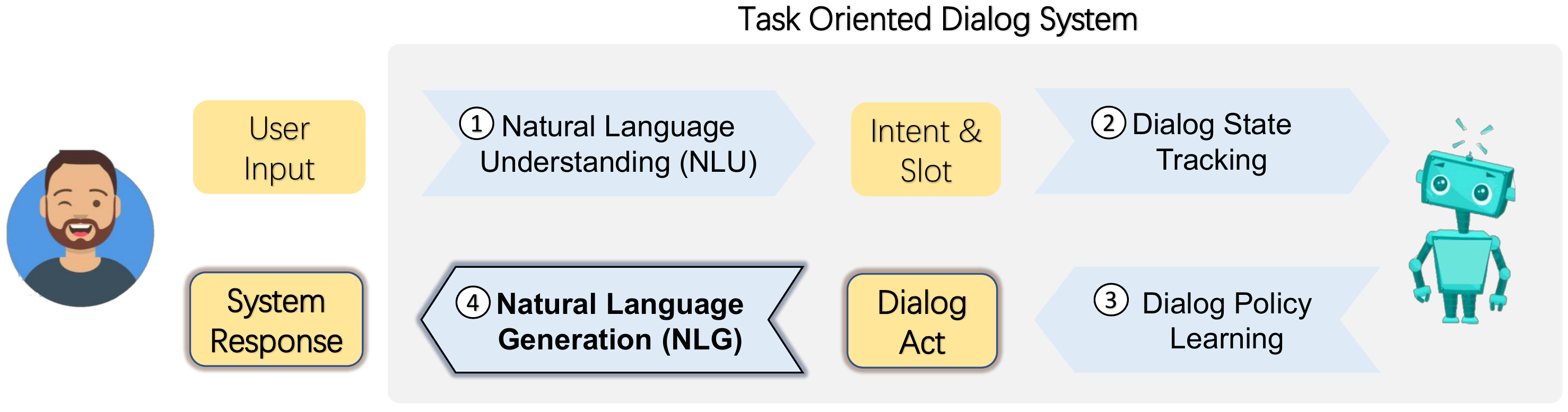}  & 
	 \hspace{-2mm}
	\includegraphics[height=3.0cm]{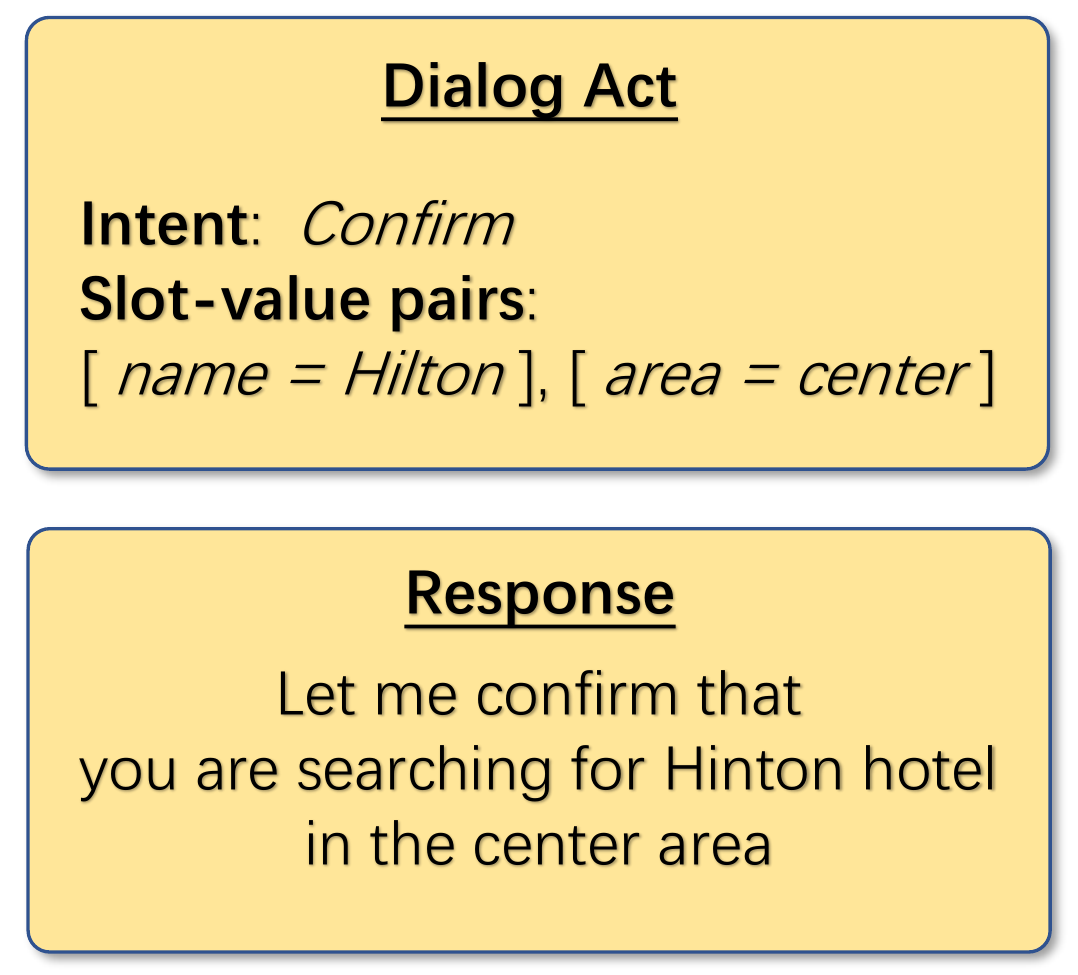} \\
	(a) The overall framework of a task-oriented dialog system \vspace{2mm} & 
	\hspace{-2mm}  (b) Dialog act \& Response \hspace{-0mm} \\ 
\end{tabular}

\caption{Illustration of the NLG module in the overall task-oriented dialog system. (a) The NLG module is highlighted with glowing black bounding boxes. (b) One example of dialog act (including intent and slot-value pairs) and its corresponding natural language response. }
\label{fig:dailog_system}
\end{figure*}

In summary, our key contributions are three-fold:
\begin{itemize}
\setlength{\itemsep}{-3pt}
    \item A new benchmark \data{} is introduced to simulate the few-shot adaptation setting where only a handful of training data from each domain is available.
    \item We propose a new model SC-GPT. To our best knowledge, this work is the first study of exploiting state-of-the-art pre-trained language models for NLG in task-oriented dialog systems.
    \item On the MultiWOZ dataset, SC-GPT creates a new SOTA, outperforming previous models by 4 points in BLEU. 
     On \data{}, SC-GPT outperforms several strong baselines such as SC-LSTM and HDSA \citep{chen-etal-2019-semantically}, showing that SC-GPT adapts to new domain much more effectively, requiring much smaller amounts of in-domain labels.
     We release our code\footnote{\url{https://github.com/pengbaolin/SC-GPT}} and dataset\footnote{Project website: \url{https://aka.ms/scgpt}} for reproducible research.
\end{itemize}


\section{Background}
A typical task-oriented spoken dialog system uses a pipeline architecture, as shown in Figure~\ref{fig:dailog_system} (a), where each dialog turn is processed using a four-step procedure.
$(\RN{1})$ Transcriptions of user’s input are first passed to the natural language understanding (NLU)
module, where the user’s intention and other key information are extracted. 
$(\RN{2})$ This information is then formatted as the input to dialog state tracking (DST), which maintains
the current state of the dialog. 
$(\RN{3})$  Outputs of DST are passed to the dialog policy module, which produces a dialog act based on the facts or entities retrieved from external resources (such as a database or a knowledge base).
$(\RN{4})$  The dialog act emitted by the dialog policy module serves as the input to
the NLG, through which a system response in natural language is generated. 
%
In this paper, we focus
on the NLG component of task-oriented dialog
systems, \ie how to produce natural language responses conditioned on dialog acts.

Specifically, {\it dialog act}
$\Acal$ is defined as the combination of intent $\Imat$ and slot-value pairs $\{(s_i, v_i)\}^P_{i=1}$:
\begin{equation}
    \Acal = [ \underbrace{~~\Imat_{~_{~}}}_{\text{Intent}}, \underbrace{(s_1, v_1), \cdots, (s_P, v_P)}_{\text{Slot-value pairs} } ]
\end{equation}
where $P$ is the number of pairs\footnote{In some literature, dialog act denotes only the type of system actions, slot-value pairs are defined as meaning representations. Throughout this paper, we follow the usage in \citet{budzianowski2018multiwoz} and use dialog acts to indicate system action and associated slot-value pairs.}, which varies in different dialog acts.   

\begin{itemize}
\setlength{\itemsep}{-3pt}
    \item {\it Intents} are usually used to distinguish different types of system actions. Typical examples include {\it inform}, {\it request}, {\it confirm}, {\it select} \etc~
    \item {\it Slot-value pairs} indicate the category and content of the information to express in the utterance, respectively.
\end{itemize}

The goal of NLG is to translate $\Acal$ into a natural language response $\xv = [x_1, \cdots, x_T]$, where $T$ is the sequence length. In Figure~\ref{fig:dailog_system} (b), we show an  example of the dialog act: 
$\textit{\texttt{confirm}~(name=Hilton, area=center)}$, and the corresponding natural language response is ``{\it Let me confirm that you are searching for Hilton in the center area}''. 
%

%



\begin{figure*}[t!]
\centering
\includegraphics[width=2\columnwidth]{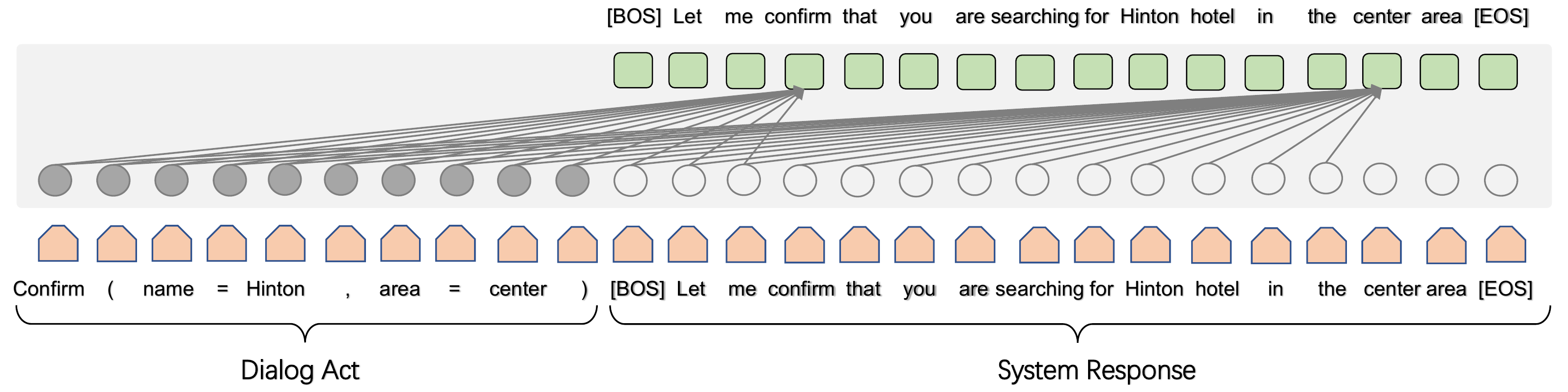}
\caption{Illustration of SC-GPT. In this example, SC-GPT generates a new word token (\eg ``\texttt{confirm}'' or ``\texttt{center}'') by attending the entire dialog act and word tokens on the left within the response.}
\label{fig:scGPT}
\end{figure*}

\section{Semantically Conditioned GPT}

We tackle this generation problem using conditional neural language models. Given training data of $N$ samples $\Dcal=\{(\Acal_n, \xv_n)\}_{n=1}^{N}$, our goal is to build a statistical model parameterized by $\thetav$ to characterize $p_{\thetav}(\xv | \Acal)$. To leverage the sequential structure of response, one may further decompose the joint probability of $\xv$ using the chain rule, casting an auto-regressive generation process as follows:
\begin{equation}
    p_{\thetav}(\xv|\Acal) = \prod_{t=1}^{T} p_{\thetav}(x_t | x_{<t}, \Acal)
    \label{eq:conditional}
\end{equation}
where $x_{<t}$ indicates all tokens before $t$.

Learning $\thetav$ is performed via maximizing the log-likelihood (MLE) of the conditional probabilities in \eqref{eq:conditional} over the entire training dataset:
\begin{equation}
    \Lcal_{\thetav}(\Dcal) =  \sum_{n=1}^{|\Dcal|} \sum_{t=1}^{T_n} \log p_{\thetav}(x_{t,n} | x_{<t,n}, \Acal_n)
\end{equation}

In this paper, we employ the Transformers \citep{transformer} to parameterize the conditionals in \eqref{eq:conditional}. To enable strong generalization and controllable ability for the learned model, we propose the following three-stage procedure as the training recipe. 

\paragraph{Massive Plain Language Pre-training.} Large models trained on massive training corpus usually generalize better to new domains. Inspired by this, we inherit the GPT-2 architecture \citep{gpt2} as the backbone language model. GPT-2 is an auto-regressive language model that leverages 12-24 layers of masked, multi-head self-attention Transformers. GPT-2 is pre-trained on extremely massive text data OpenWebText~\citep{gpt2}. It has demonstrated superior performance on characterizing human language data distribution and knowledge transfer. Given text prompts, GPT-2 can often generate realistic sentences.

\paragraph{Dialog-Act Controlled Pre-training.} To enable the guidance of dialog act in response generation, we propose to continuously pre-train the GPT-2 model on large amounts of annotated (dialog act, response) pairs. 
The pre-training dataset\footnote{The domains appearing in fine-tuning are excluded.} includes annotated training pairs from Schema-Guided Dialog corpus, MultiWOZ corpus, Frame corpus, and Facebook Multilingual Dialog Corpus. The total size of the pre-training corpus is around 400k examples.

We firstly pre-process dialog act $\Acal$ into a sequence of control codes using the following format: 
\begin{equation}
    \Acal^{\prime} = \left[ ~ \Imat~~ (~~ s_1~~ = ~~ v_1~~,~\cdots~~ s_P~ = ~~  v_P~~ )~ \right]
    \label{eq:pre_process}
\end{equation}
Meanwhile, the output sequence $\xv^{\prime}$ is pre-processed via appending $\xv$ with a special start token \texttt{[BOS]} and an end token \texttt{[EOS]}.  Finally, the sequentialized dialog act $\Acal^{\prime}$ is concatenated with its augmented response $\xv^{\prime}$, and then fed into GPT-2. During training, the prediction loss is only computed for $\xv^{\prime}$, and $\Acal^{\prime}$ provides the attended conditions. 
Since the dialog act represents the semantics of the generated sentences, we follow the naming convention of SC-LSTM, and term our model as {\it Semantically Conditioned Generative Pre-training} (SC-GPT). 
The overall architecture of SC-GPT is illustrated in Figure~\ref{fig:scGPT}.


\paragraph{Fine-tuning.} For a new domain, a dialog act usually contains novel intents or slot-value pairs, and annotated training samples are often limited. We fine-tune SC-GPT on limited amounts of domain-specific labels for adaptation. 
The fine-tuning follows the same procedure of dialog-act controlled pre-training, as described above, but uses only a few dozens of domain labels.

It is worth noticing that the above recipe has several favorable properties:

\begin{itemize}
\setlength{\itemsep}{-3pt}
    \item {\it Flexibility.} SC-GPT operates on a sequence of tokens without delexicalization, which means that SC-GPT does not assume a fixed one-hot or tree-structured dialog act representation vectors. Hence, it has great flexibility in extending to novel dialog acts. 
    \item {\it Controllability.} In contrast to GPT-2 that generates natural sentences without high-level semantic guidance, SC-GPT can generate sentences with adequate intent and slot-value information and maintain its fluency.
    \item {\it Generalizability.}  SC-GPT is able to generalize significantly better than SC-LSTM, due to the pre-training on massive plain text corpora and annotated dialog datasets.
\end{itemize}




\begin{table*}[htbp]
    \centering
    \footnotesize
    \setlength{\tabcolsep}{1.0mm}{
    \begin{tabular}{@{}p{5.5cm}@{} | @{}p{2.25cm}@{} @{}p{2.25cm}@{} @{}p{2.25cm}@{} @{}p{2.25cm}@{}  }
    \toprule
         Statistics & \textbf{E2E NLG} & \textbf{BAGEL} & \textbf{RNNLG} &   \textbf{\data{}} \\
         \midrule
         \# Domains & 1 & 1 & 4 & 7 \\
         Avg. \# Intents & 1 & 8 & 11.25 & 8.14 \\
         Avg. \# Slots & 8 & 10 & 21 & 16.15 \\
         Avg. \# Delexicalised DAs in Training & 109 &  23.9 & 794.5 & 50 \\
         Avg. \# Delexicalised DAs in Testing & 7 & 14.3 & 566.5 & 472.857 \\
         Overlap Percentage & 100\% & 99.6\% & 94.00\% & 8.82\% \\
         Avg. \# Training Instances & 42056 & 363 & 4625.5 & 50 \\
         Avg. \# Testing  Instances & 630 & 41 & 1792.5 & 472.86 \\
    \bottomrule
    \end{tabular}
    }
    \caption{Comparison of existing NLG datasets, including E2E NLG \cite{e2enlg}, BAGEL\cite{bagel}, Cambridge NLG\cite{rnnlg} and the proposed \data{}. }
    \label{tab:dataset_compare}
\end{table*}


\section{Dataset: \data{}}
\paragraph{Revisiting NLG Benchmarks.}
The three commonly used NLG datasets in developing and evaluating task-oriented dialog systems are E2E NLG~\citep{e2enlg}  BAGEL~\citep{bagel} and RNNLG~\citep{rnnlg}, as summarized in Table \ref{tab:dataset_compare}. 
We observe two issues from their shared statistics:   
$(\RN{1})$  
All the datasets contain a large number of labelled training samples for each domain, ranging from hundreds to tens of thousands. However, the cost of labeling is high in practice, \eg labeling 50 utterances is 5 hours per domain. Creating such an extensively annotated dataset for each new domain is prohibitively expensive. 
$(\RN{2})$ 
The percentage of distinct delexicalised dialog acts between training and testing data is quite small. For example, the delexicalised dialog acts in testing is 100\% covered by the training set for the E2E NLG dataset. It renders difficulties in evaluating the model's generalization ability for new domains. 

\paragraph{\data{}.}
To build a setting for more pragmatic NLG scenarios, we introduce a new dataset \data{} to better reflect real application complexity, and encourage the community to develop algorithms that are capable of generalizing with only a few domain-specific labels for each (new) domain. The dataset statistics are shown in the last column of Table \ref{tab:dataset_compare}. We see that \data{} is different from the other datasets in three aspects:
$(\RN{1})$  {\it More domains}. 
\data{} contains seven domains in total, which is larger than any existing NLG datasets. 
$(\RN{2})$ {\it Less training instances}. 
Importantly, \data{} has a much smaller number of training instances per domain, aiming to evaluate the few-shot learning ability. 
%
$(\RN{3})$ {\it Lower training/testing overlap}. \data{} has only 8.82\% overlap, significantly smaller than the other datasets, which amount to more than 90\% overlap. The average number of intents per instance in $\mathtt{Attraction}$/~$\mathtt{Taxi}$/~$\mathtt{Train}$ domain is 2, 1.33, and 2.05, respectively. In contrast, there is only one intent for each example in the other datasets.  The NLG task defined on \data{} requires the models to learn to generalize over new compositions of intents. The details of \data{} is shown in Table~\ref{tab:fewshotwoz}.

\begin{table*}[htbp]
    \centering
    \footnotesize
    \setlength{\tabcolsep}{1.0mm}{
     \begin{tabular}{@{}p{3.5cm}@{}  @{}p{2.0cm}@{}   @{}p{1.7cm}@{}  @{}p{1.7cm}@{}  @{}p{1.5cm}@{}  @{}p{2.0cm}@{}  @{}p{1.5cm}@{}  @{}p{1.5cm}@{} }
     \toprule
         Statistics & \!\!$\mathtt{Restaurant}$	
         &	$\mathtt{Hotel}$	&	$\mathtt{Laptop}$	&	$\mathtt{TV}$	
         &	\!\!$\mathtt{Attraction}$	&	$\mathtt{Taxi}$	&	$\mathtt{Train}$ \\
         \midrule
        \# Intent	&	9	&	10	&	13	&	13	&	5	&	2	&	5 \\
        \# Slot	&	21	&	19	&	22	&	22	&	10	&	7	&	13 \\
        \# DAs in training	&	50	&	50	&	50	&	50	&	50	&	40	&	50 \\
        \# DAs in testing	&	129	&	78	&	1379	&	680	&	340	&	47	&	657 \\
        Overlap Percentage	&	35.56	&	60.26	&	2.61	&	5.74	&	13.82	&	72.34	&	6.55 \\
        Avg. \#DAs per Instance	&	1	&	1	&	1	&	1	&	2	&1.33	&	2.05 \\
        \# Training Instances	&	50	&	50	&	50	&	50	&	50	&	40	&	50 \\
        \# Testing Instances	&	129	&	78	&	1379	&	680	&	340	&	47	&	657 \\
     \bottomrule
     \end{tabular}
    }
    \caption{\data{} statistics over 7 different domains. }
    \label{tab:fewshotwoz}
    \vspace{-2mm}
\end{table*}

\paragraph{Collection Protocols.} We construct \data{} via re-organizing data samples from RNNLG and MultiWOZ datasets~\citep{budzianowski2018multiwoz}. For each domain in RNNLG, we first group utterances according to their delexicalised dialog acts, and keep only one utterance as the target sentence. 
To ensure diversity, we consider three domains from MultiWOZ: $\mathtt{Attraction}$, $\mathtt{Taxi}$, and $\mathtt{Train}$. 
Since MultiWOZ is a cross-domain dataset, the dialog act of an utterance may exist in multiple domains. 
We choose to keep utterances whose dialog act appears only in one domain. 
Similar delexicalising processing is applied to ensure that each dialog act has only one target utterance. 
Finally, to simulate the few-shot learning in practice, we randomly sample 50 training examples for each domain, except the $\mathtt{Taxi}$ domain, which has 40 examples.


\section{Related Work}

\paragraph{Pre-trained Models.} 
Pre-trained language models (PLMs) have substantially advanced the state-of-the-art across a variety of natural language processing (NLP) tasks~\cite{peters2018deep,devlin2019bert,yang2019xlnet,liu2019roberta,keskar2019ctrl,raffel2019exploring}. PLMs are often trained to predict words based on their context on massive text data, and the learned models can be fine-tuned to adapt to various downstream tasks.
The closest line of research to ours are GPT-2~\cite{gpt2}, CTRL~\cite{keskar2019ctrl} and Grover~\cite{zellers2019defending}. GPT-2 first investigated missive Transformer-based auto-regressive language models with large-scale text data for pre-training. After fine-tuning, GPT-2 achieves drastic improvements on several generation tasks. One drawback of GPT-2 is the lack of high-level semantic controlling ability in language generation.
To alleviate this issue, CTRL~\citep{keskar2019ctrl} was introduced to train the model based on pre-defined codes such as text style, content description, and task-specific behavior, meanwhile Grover~\cite{zellers2019defending} was proposed to generate news articles conditioned on authors, dates \etc ~Although conceptually similar to our SC-GPT, CTRL and Grover cannot be readily applied to NLG in task-oriented dialog systems, as the conditioning codes are quite different. Another controllable generation work for GPT-2 is PPLM~\citep{dathathri2019plug}, which provides a decoding scheme to guide the generation process using key-words or classifiers, without re-training the model. 
In this paper, we focus on pre-training an NLG model conditioned on finer-grained semantic dialog acts, which are more desirable for dialog systems.


\paragraph{Dialog.} 
Various dialog systems have been developed \citep{gao2019neural}, including task-oriented dialog systems such as Rasa\footnote{https://rasa.com/}, Microsoft Bot Framework\footnote{https://dev.botframework.com/}, and Conversational Learner\footnote{https://www.microsoft.com/en-us/research/project/conversation-learner/}, and chit-chat systems such as XiaoIce~\cite{zhou2018design}, DialoGPT~\cite{zhang2019dialogpt}, Meena~\cite{adiwardana2020towards}. 
In this paper, we focus on task-oriented systems, particularly the NLG module. 
With the blooming of deep learning, neural sequential models have shown powerful capability and flexibility in NLG. Extensive efforts have been made, including new architecture choices such as RNNs~\citep{wen-etal-2015-stochastic}, attention RNNs~\citep{dusek-jurcicek-2016-sequence}, SC-LSTM  \citep{wen-etal-2015-semantically} and its variants~\citep{tran-etal-2017-neural, tran-nguyen-2017-natural}, as well as learning objectives 
\citep{zhu-etal-2019-multi}. However, they all require large amounts of annotated data to reach satisfactory performance. 
A more realistic scenario is to require much less labeling and improve the sample efficiency of models, This is especially important when deploying the models to new domains, where dialog acts need to be labelled from scratch. Our paper aims to formally set up such a research scenario by proposing a new dataset \data{}, and a new model SC-GPT.

\begin{table*}[t!]
\footnotesize
\centering
\includegraphics[height=2.5cm]{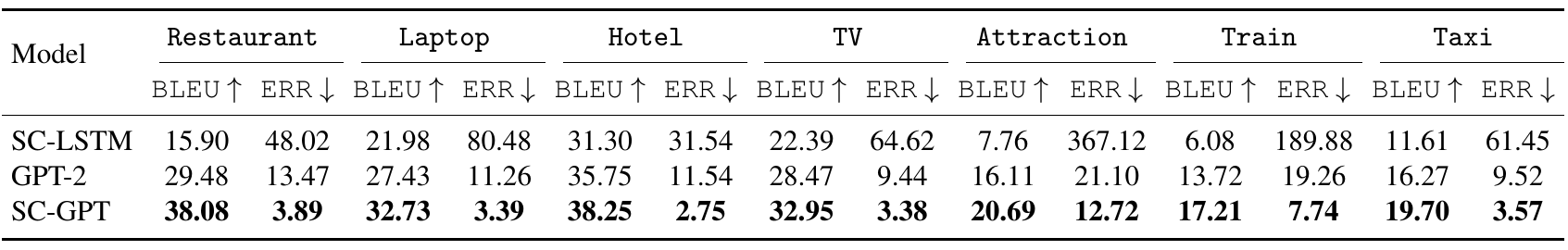}
\caption{Performance of different methods on \data{}} 
\label{tab:res_fewshotwoz}
\vspace{-2mm}
\end{table*}

\begin{table}
\footnotesize
\centering
\setlength{\tabcolsep}{1.6mm}{
\begin{tabular}{c cc} 
\toprule 
\textbf{Model} & \textbf{Informativeness} & \textbf{Naturalness} \\
\midrule
SC-LSTM & 2.29 & 2.13\\
GPT-2 & \ 2.54\textsuperscript{*}  &  \ 2.38\textsuperscript{*} \\
SC-GPT & \ \ \ 2.64\textsuperscript{*\textdagger} & \ \ \ \  2.47\textsuperscript{*\textdagger} \\
\midrule
\textit{Human} & 2.92 & 2.72 \\
\bottomrule \\[-2mm]
\multicolumn{3}{l}{\textsuperscript{*} $p < 0.005$, comparison with SC-LSTM} \\
\multicolumn{3}{l}{\textsuperscript{\textdagger} $p < 0.05$, comparison with GPT} \\
\end{tabular} }
\caption{Human evaluation on \data{}. Statistical significance is computed with a two-tailed t-test.}
\label{tab:human_eval_few}
\vspace{-2mm}
\end{table}

\section{Experiments}
In this section, we evaluate the proposed SC-GPT on the \data{} and MultiWOZ datasets to answer two research questions:
$(\RN{1})$ Is SC-GPT an effective model for strong generalization and controllability in dialog response generation?
$(\RN{2})$ Does \data{} meet the goal of effectively evaluating the generalization of NLG models in the few-shot learning setting? 

\subsection{Experimental Setup}
\paragraph{Implementation details.} The model was built upon Huggingface Pytorch Transformer \cite{Wolf2019HuggingFacesTS}. We use GPT2-Medium with 345M parameters\footnote{We also experimented using GPT2 with 117M parameters but observed significant poor performance.} as the initial checkpoint, and byte pair encodings \citep{bpe} for the tokenization. Linear rate scheduler with start rate as 5e-5 was used for both pre-training and fine-tuning. Adam~\cite{kingma2014adam} with weight decay was used to optimize the parameters. For pre-training, the model was trained with a mini-batch of 8 on an 8 Nvidia V100 machine until observing no significant progress on validation loss or up to 20 epochs, whichever is earlier. For fine-tuning on \data{}, models were trained on each domain separately with five epochs. 

\paragraph{Automatic metrics.} Following \citet{wen-etal-2015-semantically}, BLEU scores and the slot error rate (ERR) are reported. BLEU score evaluates how natural the generated utterance is compared with human readers. ERR measures the exact matching of the slot tokens in the candidate utterances. $\text{ERR}=(p+q)/M$, where $M$ is the total number of slots in the dialog act, and
$p$, $q$ is the number of missing and redundant slots in the given realisation.
For each dialog act, we generate five utterances and select the top one with the lowest ERR as the final output. 

\paragraph{Human evaluation.}
We conducted the human evaluation using Amazon Mechanical Turk to assess subjective quality. 
We recruit master level workers (who have good prior approval rates) to perform a human comparison between generated responses from two systems (which are randomly sampled from comparison systems). The workers are required to judge each utterance from 1 (bad) to 3 (good) in terms of informativeness and naturalness. \textit{Informativeness} indicates the extent to which generated utterance contains all the information specified in the dialog act. \textit{Naturalness} denotes whether the utterance is as natural as a human does. To reduce judgement bias, we distribute each question to three different workers. Finally, we collected in total of 5800 judges.

\paragraph{Baselines.} We compare with three baseline methods.
$(\RN{1})$  \textbf{SC-LSTM}~\cite{wen-etal-2015-semantically} is a canonical model and a strong baseline  that uses an additional dialog act vector and a reading gate to guide the utterance generation.
$(\RN{2})$  \textbf{GPT-2}~\cite{gpt2} is used to directly fine-tune on the domain-specific labels, without pre-training on the large-scale corpus of (dialog act, response) pairs.
$(\RN{3})$  \textbf{HDSA}~\cite{chen-etal-2019-semantically} is a state-of-the-art model on MultiWOZ. It leverages dialog act structures to enable transfer in the multi-domain setting, showing superior performance than SC-LSTM.

\subsection{\data{}}

Table \ref{tab:res_fewshotwoz} reports the automatic evaluation performance of different methods on \data{}. SC-LSTM fails to learn the generation effectively in this few-shot learning setting. The generated utterances are poor in quality and suffer from inaccurate slot rendering. 
In addition, GPT-2 performs consistently better than SC-LSTM in all the domains. It reveals the feasibility of using a pre-trained language model for NLG, though only limited annotations are available for fine-tuning. 
Importantly, SC-GPT performs significantly better than GPT and SC-LSTM in terms of both BLEU and ERR. In all the domains, SC-GPT reduces the ERR to a significantly lower level, revealing its strong controllability power. This verifies the importance of pre-training on large annotated dialog data, as SC-GPT learns how to generate utterances specified by the dialog acts accurately. 

Table \ref{tab:human_eval_few} shows the human assessment on \data{}. The results exhibit the same trend with automatic evaluation. SC-GPT outperforms GPT-2 and SC-LSTM significantly in both metrics, \ie SC-GPT can better control the generation to convey information in the dialog act while maintaining good fluency. Note that the gap between SC-GPT and human annotation is still large, indicating that the proposed \data{} exhibits an under-explored research area, and provides a large space to encourage future research for improvement.



\begin{table}
\scriptsize

\centering
\setlength{\tabcolsep}{1.6mm}{
\begin{tabular}{c c c} 
\toprule 
Model & ~~~Entity F1~~~ & ~~~BLEU~~~  \\
\midrule
SC-LSTM \cite{wen-etal-2015-semantically} & 80.42
 & 21.6 \\
HDSA \cite{chen-etal-2019-semantically} & 87.30
 & 26.48\\
GPT-2  & 87.70
 & 30.71\\
SC-GPT & \textbf{88.37}
 & \textbf{30.76}\\
\bottomrule 
\end{tabular} }
\caption{Performance on MultiWOZ}
\label{tab:res_multiwoz}
\end{table}


\begin{table}[t!]
\centering
\hspace{-2mm}
\includegraphics[height=2.6cm]{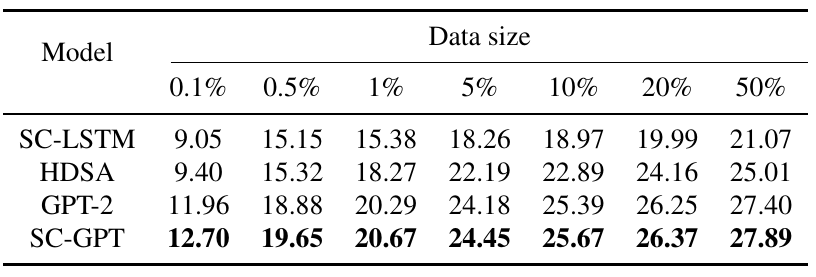}
\caption{BLEU score of different models on MultiWOZ using training data of  different sizes.}
\label{tab:res_multiwoz_pct}
\end{table}


\begin{table}[t!]
\footnotesize
\centering
\setlength{\tabcolsep}{1.6mm}{
\begin{tabular}{c cc} 
\toprule 
\textbf{Model} & \textbf{Informativeness} & \textbf{Naturalness} \\
\midrule
SC-LSTM & 2.14 & 2.33 \\
HDSA & 2.34 & 2.42 \\
SC-GPT & \ \ 2.71$^\text{*}$ & 2.69\textsuperscript{*} \\
\midrule
\textit{Human} & 2.77 & 2.61 \\
\bottomrule \\[-2mm]
\textsuperscript{*} $p < 0.005$
\end{tabular} }
\caption{Human evaluation on MultiWOZ. Statistical significance was computed with a two-tailed t-test between SC-GPT and HDSA.}
\label{tab:human_eval_multiwoz}
\end{table}

\begin{table*}[t!]\centering
\begin{minipage}{15.6cm}\vspace{0mm}    \centering
\begin{tcolorbox} 
    \centering
    \scriptsize
     \hspace{-6mm}
    \begin{tabular}{ccp{0.83\columnwidth}}
    \# & \textbf{Model} & \textbf{Generated Responses from Different Models} \\
    \midrule
    1 & \textit{\textbf{Input DA}} & Laptop\{inform(name=satellite proteus 84; type=laptop; memory=8 gb; drive=1 tb; weight=2.3 kg)\} \\[2.5pt]
    \ & \textit{\textbf{Reference}} & the satellite proteus 84 is a laptop with a 1 tb drive , 8 gb memory and weighs in at 2.3 kg \\[2.5pt]
    \ & SC-LSTM & the satellite proteus 84 is a laptop with 8 gb , with a 1 tb drive , and is \colorbox{mygreen!30}{for business computing} , and is \colorbox{mygreen!30}{for business computing} 
    ~~~\% {\it \colorbox{mygreen!30}{[businesscomputing=true]} \colorbox{blue!30}{[weight=2.3kg]} }\\[2.5pt]
    \ & GPT-2 & the satellite proteus 84 laptop has a 1 tb drive , weighs 2.3 kg and is \colorbox{mygreen!30}{used for business computing}
    ~~~\% {\it \colorbox{mygreen!30}{[businesscomputing=true]} \colorbox{blue!30}{[memory=8 gb]} }\\[2.5pt]
    \ & SC-GPT & the satellite proteus 84 is a laptop with 8 gb memory , 1 tb drive , and a weight of 2.3 kg \\[2.5pt]
    \midrule
    2 & \textit{\textbf{Input DA}} & Restaurant\{inform\_only\_match(name=marnee thai and thanh long; pricerange=moderate; near=outer sunset; goodformeal=dinner)\} \\[2.5pt]
    \ & \textit{\textbf{Reference}} & marnee thai and thanh long are the only 2 moderately priced restaurants near outer sunset that are good for dinner  \\[2.5pt]
    \ & SC-LSTM & marnee thai and thanh long is is moderate restaurant restaurant and good good good dinner 
    ~~~~~\% {\it \colorbox{blue!30}{[near=outer sunset]} \colorbox{blue!30}{[inform\_only\_match]} } \\[2.5pt]
    \ & GPT-2 & there is a moderately priced restaurant called marnee thai and thanh long that is near the outer sunset area 
    ~~~\% {\it \colorbox{blue!30}{[goodformeal=dinner]} \colorbox{blue!30}{[inform\_only\_match]}  }\\ [2.5pt]
    \ & SC-GPT & marnee thai and thanh long is the only restaurant that serves moderately priced food near outer sunset and good for dinner \\[2.5pt]
    \midrule
    3 & \textit{\textbf{Input DA}} & Train\{inform(time=50 minutes)\}, hotel\{request(stars=?; area=?), inform(choice=5)\} \\[2.5pt]
    \ & \textit{\textbf{Reference}} &  travel time is 50 minutes . i found 5 hotels you may like . do you have a star rating request or an area you prefer \\[2.5pt]
    \ & SC-LSTM &  there are \colorbox{red!30}{5 trains} that are 50 minutes . do you have a preferred \colorbox{red!30}{departure location} ? \\ 
    \ & \ & \% {\it \colorbox{red!30}{[train\{inform(choice=5),request(departure=?)\}]}\colorbox{blue!30}{[hotel\{request(stars=?; area=?), inform(choice=5)\}]} } \\
    \ & HDSA & there are \colorbox{red!30}{5 trains} that meet your criteria . do you have a preference on the area or star rating ? 
    ~~\% {\it \colorbox{red!30}{[train\{inform(choice=5,} \colorbox{blue!30}{time=50 minutes)\}]}  \colorbox{blue!30}{[hotel\{inform(choice=5)\}]} } \\ [2.5pt]
    \ & SC-GPT &  there are 5 hotels that meet your criteria . the trip will last 50 minutes . do you have an area preference or star rating you would like ? \\[2.5pt] \vspace{-3mm}
    \end{tabular}
\end{tcolorbox}
\vspace{-2mm}
\caption{Examples of generated utterances from different models, along with its corresponding dialog acts (DAs) and references. The first two examples are sampled from \data{} and the last one is from MultiWOZ. Each generated utterance is followed by a brief description explaining the errors (starting with ``\%''). (Better viewed in color. \colorbox{red!30}{wrong}, \colorbox{mygreen!30} {redundant}, \colorbox{blue!30} {missing} information)}
    \label{tab:examples}
\end{minipage}
\end{table*}

\subsection{MultiWOZ}
The results on MultiWOZ are shown in Table \ref{tab:res_multiwoz}. 
Following~\citet{chen-etal-2019-semantically}, Entity F1~\cite{wen2016network} is used to evaluate the entity coverage accuracy (including all slot values, days, numbers, and reference, \etc).
Again, SC-GPT achieves the best performance on BLEU score. Note that GPT-2 performs similarly with SC-GPT on the full MultiWOZ dataset, this is because MultiWOZ contains 57k utterances, which is large enough for GPT-2 to achieve good performance. 
The results also confirm that with enough annotated data, conditional language model formulation performs significantly better than HDSA, a strong competitor that leverages graph/tree-structure information to encode dialog acts. 

To study how SC-GPT performs with different training data sizes. We further conduct experiments with varying percentages of training data on MultiWOZ, ranging from 0.1\% (50 examples) to 50\%. As shown in Table \ref{tab:res_multiwoz_pct}, the observations are consistent with \data{}. SC-GPT performs consistently better than GPT-2, HDSA, and SC-LSTM for a wide range of dataset sizes, and the improvement is more substantial when the fewer numbers of in-domain labels are used for fine-tuning.

Table \ref{tab:human_eval_multiwoz} shows the human assessment results on MultiWOZ. The results are consistent with the automatic evaluation. It is interesting to see that 
$(\RN{1})$ the gap between the new state-of-the-art method (\ie SC-GPT ) and human performance on \data{} (as shown in Table~\ref{tab:human_eval_few}) is much larger than that on MultiWOZ; 
$(\RN{2})$ the human rating on the naturalness of SC-GPT is even higher than humans on MultiWOZ, while there is a visible gap on \data{}. These results demonstrate that \data{} presents a challenging few-shot learning setting, SG-GPT serves as a simple and strong baseline in this setting, and the combined provides a platform for researchers to develop NLG models that are able to generalize to new domains and generate semantically conditioned and fluent responses.

\subsection{Analysis}
\begin{table}[t!]
\footnotesize
\centering
\setlength{\tabcolsep}{1.0mm}{
\begin{tabular}{l cc cc } 
\toprule 
\multirow{2}{*}{Model} &
\multicolumn{2}{c}{$\mathtt{Seen}$} &
\multicolumn{2}{c}{$\mathtt{Unseen}$} \\

\cmidrule(l){2-3} \cmidrule(l){4-5} 

&$\texttt{BLEU}$ $\uparrow$ & $\texttt{ERR}$ $\downarrow$ 
&$\texttt{BLEU}$ $\uparrow$ & $\texttt{ERR}$ $\downarrow$\\
\midrule
SC-LSTM & 
23.05 & 40.82 & 12.83 & 51.98 \\
GPT-2 &
30.43 & 3.26 & 27.92 & 17.36 \\
SC-GPT  &
 \textbf{40.28} & \textbf{1.09} & 
\textbf{36.69} & \textbf{4.96} \\
\bottomrule 
\end{tabular}
}
\caption{Performance of different methods on seen DAs and unseen DAs in restaurant domain.} 
\label{tab:res_seen_unseen}
\end{table}

We perform detailed analysis to investigate  SG-GPT's 
\textit{flexibility}, \textit{controllability} and \textit{generalizability}. 
The test set is split into two subsets - \textit{seen} and \textit{unseen}. If a dialog act of an example appears in the training set, the example is marked as \textit{seen}; otherwise, it is marked as \textit{unseen}. Table \ref{tab:res_seen_unseen} compares different models on the seen and unseen subsets in the $\mathtt{restaurant}$ domain. SC-GPT yields higher BLEU and lower ERR, and the improvement is more significant on the unseen set. 
For example, SC-GPT reduces ERR to 4.96, an order of magnitude lower than SC-LSTM and only 1/3 of GPT-2. This demonstrates that SC-GPT generalizes well to novel dialog acts, and is able to precisely ground in them to compose fluent responses.
This is further confirmed by the quantitative comparison in Table \ref{tab:examples},  where we compare the generated utterance examples of different models. While the baseline methods prone to over-generate or miss important slots, SC-GPT can successfully generate fluent natural language utterances that share precise semantic conditions with the ground-truth references.

%


\begin{table*}[t!]\centering
\begin{minipage}{15.6cm}\vspace{0mm}    \centering
\begin{tcolorbox} 
    \centering
    \scriptsize
     \hspace{-6mm}
    \begin{tabular}{ccp{0.83\columnwidth}}
    \# & \textbf{Model} & \textbf{Generated Responses from Different Models} \\
    \midrule
    \ & \textit{\textbf{Original DA}} & ack\_makereservation ( price = moderate ; restaurantcusine = chinese ; location = center ; peoplecount = 2 ) \\[2.5pt]
    \ & \textit{\textbf{Reference}} & ok making a reservation for restaurant moderate chinese in center for two people? \\[2.5pt]
    \midrule
    1 & \textit{\textbf{Input DA}} & ack\_makereservation ( price = moderate ; restaurantcusine = \colorbox{blue!30}{japanese} ; location = \colorbox{blue!30}{bellevue square}; peoplecount = 2 ) \\[2.5pt]
    \ & GPT-2 & ok making reservation for moderate japanese restaurant in bellevue square for 2 people? \\[2.5pt]
    \ & SC-GPT & ok making a reservation for a moderate japanese cuisine restaurant in bellevue square for 2 people? \\[2.5pt]
    \midrule
    2 & \textit{\textbf{Input DA}} & ack\_makereservation ( \colorbox{mygreen!30}{restaurantname = sakura} ; price = moderate ; restaurantcusine = \colorbox{blue!30}{japanese} ; location = \colorbox{blue!30}{bellevue sqare}; peoplecount = 2 ) \\[2.5pt]
    \ & GPT-2 & ok making a reservation for 2 people sakura moderate japanese restaurant located bellevue sqare for 10 minutes? \\ [2.5pt]
    \ & SC-GPT & ok making a reservation for sakura, moderate japanese cuisine in bellevue square for 2 people? \\[2.5pt]
    \midrule
    3 & \textit{\textbf{Input DA}} & ack\_makereservation ( price = \colorbox{blue!30}{expensive} ; restaurantcusine = japanese ; \colorbox{red!30}{location = center} ; \colorbox{red!30}{peoplecount = 2})
 \\[2.5pt]
    \ & GPT-2 & ok making reservation for japanese cuisine expensive? \\ [2.5pt]
    \ & SC-GPT &  ok making reservation for expensive japanese cuisine? \\[2.5pt]
        \midrule
    \# & \textbf{Model} & \textbf{Generated Responses from Different Models} \\
    \midrule
        \ & \textit{\textbf{Original DA}} & ack\_makereservation ( childrenfriendly = true ) \\[2.5pt]
    \ & \textit{\textbf{Reference}} & make reservation for a children friendly restaurant ? \\[2.5pt]
    \midrule
    1 & \textit{\textbf{Input DA}} &  ack\_makereservation ( \colorbox{mygreen!30}{restaurantname = sakura } ; childrenfriendly = true ; \colorbox{mygreen!30}{price = moderate} ) \\[2.5pt]
    \ & GPT-2 & make reservation at sakura child friendly restaurant and make table moderate price? \\ [2.5pt]
    \ & SC-GPT &  make reservation for restaurant sakura moderate price and children friendly restaurant? \\[2.5pt] \vspace{-3mm}
    \end{tabular}
\end{tcolorbox}
\vspace{-2mm}
\caption{Examples of generated utterances with novel dialog acts. SC-GPT produces better utterances than GPT-2 for with edited dialog acts. Since both models produce similar responses to references for the original dialog act, the results are not shown here.  (Better viewed in color. \colorbox{mygreen!30}{insert a slot}, \colorbox{blue!30}{substitute a slot value}, \colorbox{red!30}{ delete a slot}).}
    \label{tab:new_domain_examples}
\end{minipage}
\vspace{-2mm}
\end{table*}

We further simulate the process when deploying SC-GPT for a new domain, using the examples provided in the RASA dialog toolkit \footnote{https://github.com/RasaHQ/rasa/tree/master\\/examples/restaurantbot}. 
We first fine-tune SC-GPT using a few training examples (only 16 instances in this new domain), and then generate utterances based on novel dialog acts that are unseen in training data, shown in Table \ref{tab:new_domain_examples}.
In practice, it is desirable for an NLG system to deal with an extending domain whose dialog acts change dynamically. We simulate the setting by editing the original input dialog acts, such as inserting or deleting a slot, or substituting a slot value.

Since SC-LSTM is infeasible in the setting of an extending domain, we compare SC-GPT with GPT-2.
Results show that SC-GPT produces better utterances than GPT-2. 
SC-GPT can generate reasonably good natural language responses with different combinations of editing operations, showing its high flexibility to generalize to new dialog acts with very limited training data, and produce controllable responses.
\vspace{-2mm}
\section{Conclusion and Future Work}
\vspace{-2mm}
In this paper, we have made two major contributions towards developing a more pragmatic NLG module for task-oriented dialog systems:
$(\RN{1})$ 
A new benchmark \data{} is introduced to simulate the few-shot learning scenarios with scarce labelled data in real-world applications. 
$(\RN{2})$ 
A new model SC-GPT is proposed to endow the NLG module with strong semantically controlling and generalization ability.
Empirical results on both \data{} and MultiWOZ show that SC-GPT achieves the best overall performance in both automatic and human evaluations.

There are two interesting directions for future work. The first is to design mechanisms to generate more interpersonal responses which are proven to help improve user experiences \citep{li2016diversity,zhou2018design}. 
The other is to generalize the generative pre-training idea to all four modules in the dialog system pipeline for end-to-end training. Since these four modules process information in order, one may organize their input/output as segments, and pre-train a segment-level auto-regressive model.


\bibliography{acl2020}
\bibliographystyle{acl_natbib}

\appendix

\end{document}